%% file: template.tex
\documentclass{article}

\usepackage{arxiv}
\usepackage{fancyvrb}
\usepackage[utf8]{inputenc} 
\usepackage[T1]{fontenc}    
\usepackage{hyperref}       
\usepackage{url}            
\usepackage{booktabs}       
\usepackage{amsfonts}       
\usepackage{nicefrac}       
\usepackage{microtype}      
\usepackage{lipsum}		
\usepackage{graphicx}
\usepackage{natbib}
\usepackage{doi}
\usepackage[frozencache=true,cachedir=minted-cache]{minted}

\title{AutoScrum\texttrademark: Automating Project Planning Using Language Model Programs}

\date{June 4, 2023}	

\author{ \href{https://orcid.org/0009-0000-3718-6337}
	{\includegraphics[scale=0.06]{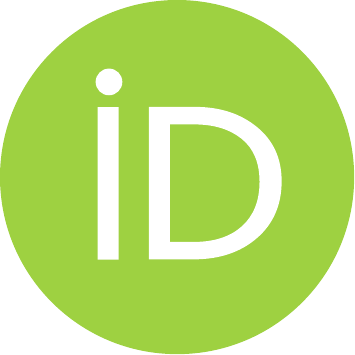}\hspace{1mm}Martin ~Schröder}\thanks{} \\
	\href{https://www.linkedin.com/in/martinschroder/}{LinkedIn: martinschroder} \\
	Swedish Embedded Consulting Group Research\\
	\texttt{martin.schroder@swedishembedded.com} \\
}

\renewcommand{\undertitle}{Leveraging language model reasoning in agile planning
and execution}

\hypersetup{
pdftitle={AutoScrum\texttrademark: Automating Project Planning Using Large Language Model
Programs},
pdfsubject={q-bio.NC, q-bio.QM},
pdfauthor={Martin ~Schröder},
pdfkeywords={LLMs, ChatGPT, Autonomous Agents},
}

\begin{document}
\maketitle

\begin{abstract}
\input{parts/abstract}

\end{abstract}


\section{Introduction}
\input{parts/introduction}

\section{Related Work}
\input{parts/related-work}

\section{Scrum Overview}
\input{parts/scrum}

\section{Methodology}
\input{parts/method-overview}

\subsection{Requirement Identification}
\input{parts/method-requirements}

\subsection{Product Feature Extraction}
\input{parts/method-product-features}

\subsection{User Story Mapping}
\input{parts/method-story-mapping}

\subsection{Acceptance Criteria}
\input{parts/method-acceptance}

\subsection{Task Decomposition}
\input{parts/method-task-decomposition}

\subsection{Task Completion}
\input{parts/method-task-completion}

\subsection{Shortcut Planning}
\input{parts/method-shortcut-plan}

\section{Experimental Results}
\input{parts/results-overview}

\subsection{Current To Desired Situation}
\input{parts/results-scrum}

\subsection{Shortcut Planner Results}
\input{parts/results-shortcut}

\subsection{Effect Of Temperature}
\input{parts/results-temperature}

\section{Final Thoughts}
\input{parts/final-thoughts}

\bibliographystyle{unsrtnat}
\bibliography{template}

\end{document}

%% file: parts/abstract.tex
Recent advancements in the field of large language models have made it possible
to use language models for advanced reasoning. In this paper we leverage this
ability for designing complex project plans based only on knowing the current
state and the desired state. Two approaches are demonstrated - a scrum based
approach and a shortcut plan approach. The scrum based approach executes an
automated process of requirements gathering, user story mapping, feature
identification, task decomposition and finally generates questions and search
terms for seeking out domain specific information to assist with task
completion. The shortcut approach looks at most recent snapshot of the current
and desired state and generates the next most reasonable task to do in order to
get to the desired state as quickly as possible. In this paper we automate
everything using a novel concept of "Language Programs". These are programs
written in natural language designed to process input data through the language
model. Guidance language is used for all LLM programs. All demo source code for
this paper is available at \url{https: //github.com/autoscrum/autoscrum}

%% file: parts/introduction.tex
The exponential growth of large language models in recent years has seen
its influence permeate various sectors and industries, fostering an era of
unprecedented growth in productivity. Models such as GPT-4 are now capable of
understanding intricate logic and complete reasoning tasks that go far beyond
simple comprehension of language.

This paper takes the concept of Autonomous Agents best described in the
"Voyager" paper (\cite{wang2023voyager}) several steps further and uses language
models to greatly simplify agile project planning by using these models to help
with scrum based project execution.

The process is designed to take you from a well defined current state to a well
defined desired state. We develop an innovative process for bridging this gap
using large language models.

\noindent\makebox[\textwidth]{\includegraphics[width=\textwidth]
{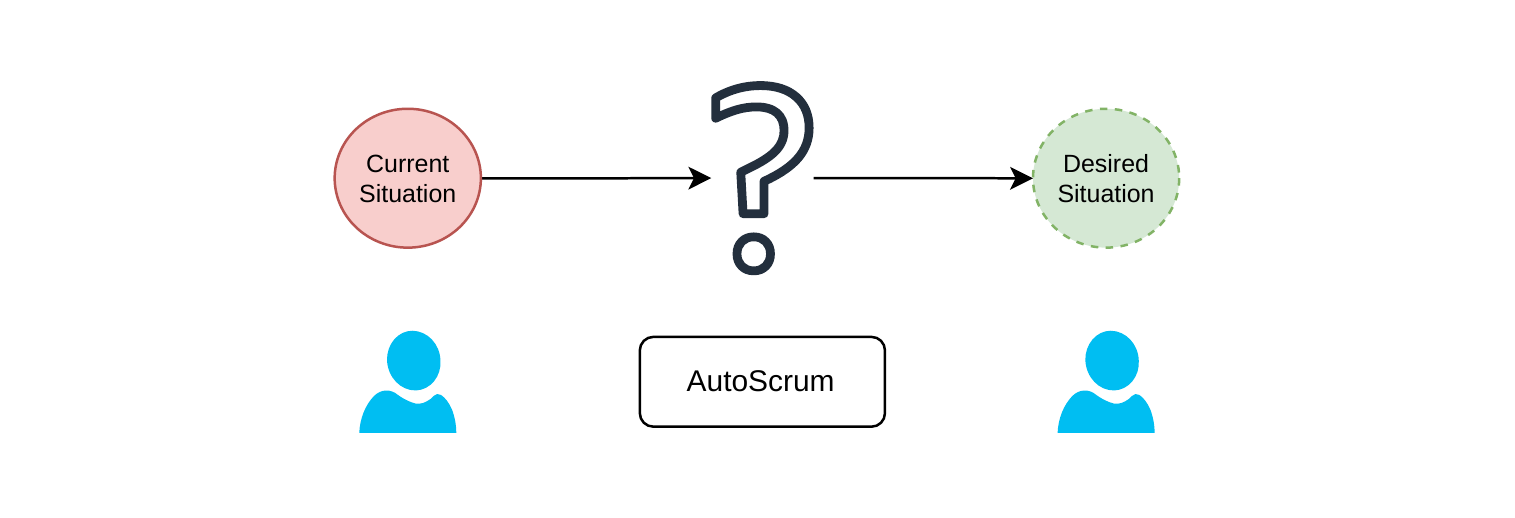}}

We use GPT-3.5-turbo model with Guidance language as a way to write and execute
natural language programs on GPT-3.5 to accomplish the tasks of scrum project
planning. GPT-4 can be used as well, and would provide vastly better results,
but since GPT-3.5 is already quite capable and a much more economical
alternative, we opt for using GPT-3.5 throughout this paper.

This paper also covers a "short-cut" approach where we also start with the
knowledge of where we are and where we want to go (expressed as any measurable
quantity) and we then use the language model iteratively to suggest the next
task we should do. This approach is conceptually very similar to the flood-fill
maze solving algorithm where at every iteration, we reevaluate the situation
using the language model and the newly uncovered information about our current
state and then try to find the shortest path to the goal using the sum total of
information available.

Full python source code for this paper is presented in the end of the paper and
is also available on GitHub:

\begin{center}
	\url{https://github.com/autoscrum/autoscrum}
\end{center}

All code in this paper uses Microsoft Guidance syntax which is an open source
project that implements a framework for advanced prompting. Guidance allows us
to write "programs" that we can run on a variety of language models, including
open source language models and ChatGPT.

\begin{center}
	\url{https://github.com/microsoft/guidance}
\end{center}

%% file: parts/related-work.tex
In the paper "Tool Manipulation Capability of Open Source Language Models"
(\cite{xu2023tool}) the authors greatly improve the accuracy of open source
models by adding API usage examples. The sweet spot where all mentioned models
including ChatGPT get the best results is with 4 examples.

In the paper "Sparks of Artificial General Intelligence: Early experiments with
GPT-4" ( \cite{bubeck2023sparks}) we can clearly see that the GPT-4 model is
much more than just a model of the language. GPT-4 displays properties that are
better described as being a reasoning engine rather than just a language model.

This reasoning capacity has been demonstrated by actual results in software
packages like AutoGPT and in a paper titled "Voyager: An Open-Ended Embodied
Agent with Large Language Models" (\cite{wang2023voyager}) where the authors
developed a novel concept of a skill library where new skills could be developed,
debugged and stored by a GPT-4 agent and then used to solve problems in a
Minecraft world using observations of the environment translated into text.

Another interesting work in ChatGPT-3/4 performance on biometrical data by
storing an embeddings database of examples which are fed to ChatGPT to improve
its accuracy titled "Improving accuracy of GPT-3/4 results on biomedical data
using a retrieval-augmented language model" (\cite{soong2023improving}).

In "Plan-and-Solve Prompting" (\cite{wang2023planandsolve}) the authors
demonstrate how detailed instructions given to the language model can improve
the ability of the model to solve complex tasks.

Attempts have also been made at connecting language models with memory systems
such as "Think Before You Act" (\cite{kang2023think}). In such scenarios
language models are used to transform a piece of content into a vector embedding
and then search vector embeddings by similarity. The paper OlaGPT
(\cite{xie2023olagpt}) pairs the model with multiple types of memory libraries
like tools library, thought library and a facts library.

In the paper "Ghost in the Minecraft: Generally Capable Agents for Open-World
Environments" \cite{zhu2023ghost} the authors discuss a novel framework
integrates Large Language Models (LLMs) with text-based knowledge and memory,
aiming to create Generally Capable Agents (GCAs) in Minecraft.

In the "InternGPT: Solving Vision-Centric Tasks by Interacting with ChatGPT
beyond language" \cite{liu2023interngpt} the authors introduce a framework
that integrates chatbots that have planning and reasoning capabilities, such as
ChatGPT, with non-verbal instructions like pointing movements that enable users
to directly manipulate images or videos on the screen.

%% file: parts/scrum.tex
As the first step in our study we are going to clarify the whole scrum process
to make sure that it is neatly laid out in front of us so that we have a good
understanding of what it is all about.

This is a simplified version of the whole Scrum process but I think it captures
everything that we are interested in and it also provides us with the framework
needed for automation.

There are a few important processes in Scrum:
\begin{itemize}
	\item \textbf{Backlog Creation}: this process is executed by product owners
		and stakeholders to identify the initial set of requirements, product
		features and user stories. This process is the first step that fills the
		backlog with a rough idea of what will need to be done.
	\item \textbf{Backlog Refinement}: this process is executed at regular
		intervals to further refine the backlog by eliminating stories that are
		no longer relevant, breaking down large user stories into smaller ones,
		estimating effort and re-prioritizing user stories.
	\item \textbf{Task Decomposition}: this process is focused on user story
		refinement and at this step the team decomposes a user story into tasks
		that the team thinks will be needed to execute the user story.
	\item \textbf{Scrum Execution}: this process is executed by the team "in the
		field". The team takes the refined backlog, breaks it down into sprints
		(2 week long increments) and executes batches of tasks in parallel among
		team members.
\end{itemize}

\subsection{Backlog Creation}
\input{parts/scrum-backlog-creation}

\subsection{Backlog Refinement}
\input{parts/scrum-backlog-refinement}

\subsection{Task Decomposition}
\input{parts/scrum-task-decomposition}

\subsection{Scrum Execution}
\input{parts/scrum-execution}

%% file: parts/scrum-backlog-creation.tex
The pseudo-code for backlog refinement can be expressed as follows:

\begin{Verbatim}[tabsize=2]
Initialize empty product backlog
Initialize empty set of product features
Initialize empty set of product goals

While product is not complete:
  Gather requirements from stakeholders:
    Identify the needs and expectations of the stakeholders

  For each requirement:
    Define a product feature that satisfies the requirement
    Add the feature to the set of product features
    Define a product goal that is achieved by the feature
    Add the goal to the set of product goals

    Define a user story for the feature:
      The user story should describe who the user is, what they want, and why they want it
      Add the user story to the product backlog
      Define acceptance criteria for the user story:
        The acceptance criteria should clearly define when the story is considered "done"
        Attach the acceptance criteria to the user story
  Prioritize the user stories in the product backlog based on the business
  value, risk, difficulty, and necessity
End While
\end{Verbatim}

%% file: parts/scrum-backlog-refinement.tex
The backlog also needs to be refined regularly and this is done by product
owners and project managers. This process can happen in parallel to the sprint
execution process but it does not (should not) affect sprint execution since a
sprint always executes a preplanned agenda (a collection of tasks in the sprint
backlog).

\begin{Verbatim}[tabsize=2]
Initialize product backlog with user stories
Initialize empty refined product backlog 

While product is not complete:
  For each user story in the product backlog:
    If user story is not relevant anymore:
      Remove user story from product backlog
    Else:
      If user story is too large:
        Break down user story into smaller user stories
      If user story is not estimated:
        Estimate effort to implement user story
      If user story is not clear:
        Refine the description and acceptance criteria of the user story
      Re-prioritize user story based on business value, risk, difficulty, and necessity
  Move refined user stories to the refined product backlog
End While
\end{Verbatim}

When a new sprint is about to start, the topmost user stories from the refined
product backlog are selected for the new sprint, considering the team's
capacity.

%% file: parts/scrum-task-decomposition.tex
The process of decomposing user stories into tasks can be expressed as the
following pseudo-code:

\begin{Verbatim}[tabsize=2]
While product is not complete:
  For each user story in the product backlog:
    Initialize an empty set of tasks for the user story
    While the user story is not sufficiently decomposed into tasks:
      Identify a task that is part of the user story
      Ensure the task is small and manageable,
        ideally can be done in a few hours to a day by a team member
      Add the task to the set of tasks for the user story
    Attach the set of tasks to the user story
End While
\end{Verbatim}

%% file: parts/scrum-execution.tex
Once the team is happy with the tasks that have been identified and put into the
product backlog, the team plans and executes the work in sprints. Sprint
planning is an opportunity to further refine the backlog, estimate effort given
current knowledge of the progress and pack work into a sprint which can be
executed by a team.

The biggest value of this process is that the whole team knows at the start
exactly what they should be doing and they regularly check in with each other
(daily stand-up) to make sure that their ideas do not diverge.

The scrum backlog execution process, including sprint planning, can be expressed
in pseudo-code as follows:

\begin{Verbatim}[tabsize=2]
Initialize empty sprint backlog
While product is not complete:
  Plan a new sprint:
    Determine sprint length (commonly two weeks)
    From the product backlog, select the highest priority user stories up to the team's capacity
    For each user story, break it down into tasks
    Move these user stories to the sprint backlog
    Estimate effort for each task (often using story points)
  For each day in the sprint:
    Hold daily stand-up meeting:
      Each team member answers three questions:
        1. What did I complete since the last meeting?
        2. What will I work on before the next meeting?
        3. Do I see any impediment that could prevent me or the team from meeting our goals?
    Update sprint backlog to reflect current status of tasks
    Team members work on tasks:
      If a task is complete, mark it as done
      If an issue is found, add it to the sprint backlog
  At the end of the sprint:
    Hold sprint review meeting:
      Present what was completed during the sprint to stakeholders
      Update the product backlog if new changes are needed based on feedback
    Hold sprint retrospective meeting:
      Discuss what went well during the sprint
      Discuss what could be improved in the next sprint
      Make a plan for implementing improvements in the next sprint
    Move any incomplete stories back into product backlog for next sprint plan
\end{Verbatim}

%% file: parts/method-overview.tex
In this section we are going to identify the elements of the Scrum process that
we can automate using language models and define a methodology for integrating
large language models with the scrum process.

It is not necessary to use GPT-3 or GPT-4, but these models are the most
advanced and provide very good results. However, the guidance language programs
that we are using in this paper can be executed without any changes on any
language model that is supported by guidance. Although the results may be
suboptimal compared to ChatGPT.

Once we have identified what we need to automate, we are going to write Guidance
programs for each task and then integrate the whole process together using
Python.

\subsection{Identifying Automation Tasks}
We can easily identify the following tasks from the pseudo code that we can
automate using language models.

Note that all of these tasks will be done by pairing the language model with the
developer team. You simply don't want to rely completely on the language model.
Instead, the language model can brainstorm and suggest the next steps and then
you let the team filter the results of the language model as the process
progresses at each step.

The process is highly iterative where we will be generating small chunks at a
time, reviewing them and then committing them into the database.

\begin{itemize}
	\item \textbf{Requirement Identification}: we identify
		requirements by looking at the discrepancy between our niche
		participants' current state and their desired state using the language
		model.
	\item \textbf{Product Feature Identification}: Defining product features
		based on a list of requirements. This will take as input the list of
		requirements and then generate product features.
	\item \textbf{User Story Mapping}: Breaking down product features into user
		stories based on identified product features and the transition between
		user's current state and their desired state.
	\item \textbf{Acceptance Criteria}: Defining acceptance criteria for each
		user story which can be automatically verified.
	\item \textbf{Task Decomposition}: We take our user stories and break them
		down into tasks and subtasks based on the identified acceptance criteria
		and other parameters of each story.
	\item \textbf{Task Completion Assistance}: Using the information we have
		about each task we can automatically suggest search queries and
		questions to the user to help them find the right information for
		completing the task. This can be augmented using an internal knowledge
		database from which we can extract information relevant for completing
		the task.
	\item \textbf{Shortcut Planning}: This is a task completion algorithm that
		can be used to complete an acceptance criteria by iteratively suggesting
		the next step to take based on the current and the desired situations.
		It uses the reasoning capability of the language model to suggest the
		most urgent next task.
\end{itemize}

The autoscrum process itself looks like this. The green icons represent parts of
the process that we can augment using language models.

\noindent\makebox[\textwidth]{\includegraphics[width=\textwidth]
{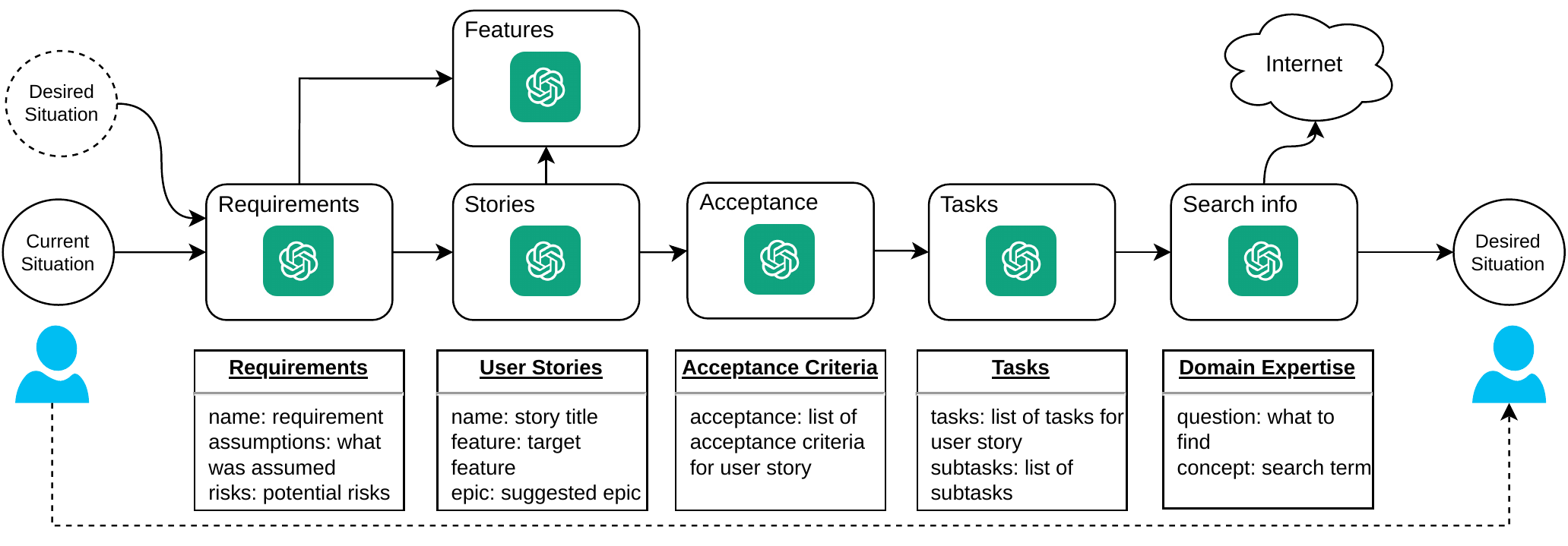}}

The shortcut planning approach uses a single language program that suggests next
most urgent task based on tasks completed so far and the difference between
current and desired situations. This language program is executed iteratively
until the desired state is reached.

\noindent\makebox[\textwidth]{\includegraphics[width=\textwidth]
{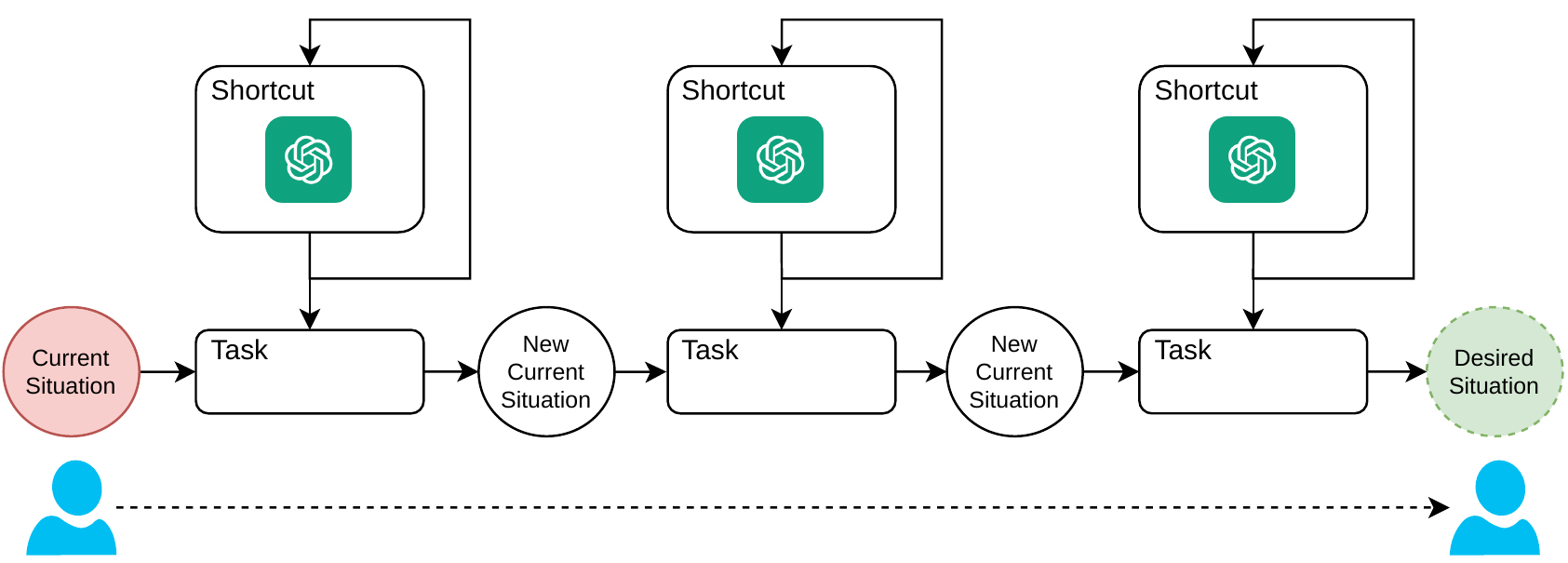}}

AutoScrum works by taking into account limited information about the project you
are working on heavily automating the process of breaking down the whole project
into manageable and achievable tasks. If you have ever worked in Scrum projects
you know that this process is typically very time consuming and it is often
difficult to do it properly because there is a lot of information to take into
account. Language models are excellent at directing your attention to the topics
that matter and this paper shows you how to leverage this ability to simplify
the process.

%% file: parts/method-requirements.tex
The automatic requirement identification process works by using the language
model to analyze the discrepancy between a customer's current state and a
desired state.

A "product" is just a vehicle that takes the customer from their current state
to their desired state. Every product on the market, is either explicitly or
implicitly taking the buyer from a certain current state to a certain desired
state. A product is a bridge between these two states.

To identify the requirements of our product we use the language model to
generate product requirements based on the transformation that we want to
achieve for the customer. We are using the language model to identify these
requirements solely from the discrepancy between the current and desired states.

\begin{itemize}
	\item \textbf{Niche}: a subset of customers that have similar features.
		Grouped by similarity. In this paper this is a hyperparameter we must
		define ourselves. Who are we targeting?
	\item \textbf{Current Situation}: each member in the niche would have a
		similar current situation (otherwise they would not need the product).
		This is another hyperparameter that we must identify ourselves through
		customer knowledge.
	\item \textbf{Desired Situation}: this is the desired situation that the
		product we are designing is meant to help the customer achieve.
\end{itemize}

Requirement identification uses a Guidance language program that is designed to
solve for the shortest path between the current situation and the desired
situations.

\begin{Verbatim}
{{~! Generation of user requirements from current and desired state }}
{{#system~}}
You are a helpful assistant that helps me to identify product requirements for a product.

Your responses contain only valid JSON.

Your task today is to generate product requirements that can take the customer from their current
situation to their desired situation.

I will provide you with the following information:
Product: name of the product we are building
Niche: aggregation of people and companies for whom this product is made
Current Situation: the current situation of the customer when they buy the product
Desired Situation: the desired situation of the customer that the product is designed to achieve
Existing Requirements: a list of existing product requirements that you should not mention again

Make sure that:
1) Each requirement is highly relevant for moving the customer from their current state to their
   desired state.
2) You should stop generating requirements when you think that the list is complete.
3) The full list of existing requirements and the new requirements that you will generate together
   should represent a complete set of requirements that can then be used to identify product
   features in scrum process.
4) Generate only at most {{count}} requirements at a time or an empty list if you are done.

The requirements should be returned as a JSON list in the following format.

RESPONSE FORMAT:
[
  {
    "name": "name of the requirement",
    "description": "a description that accurately captures the essence of the requirement",
    "customer_type": "the type of the end user",
    "stakeholders": "if customer is a company then write out roles of people for whom this is
    important. If customer is an individual then just write 'user'"
    "importance": number between 1 and 5 signifying importance (1 least, 5 most important)
    "assumptions": "assumptions that you have made when considering this requirement important",
    "risks": "possible risks associated with this requirement",
  },
  ...
]

Current state and desired state will be represented by multiple variables where you have to generate
requirements that optimize the transition from current state to the desired state.
{{~/system}}
{{#user~}}
Product: {{product}}
Niche: {{niche}}
Current Situation: {{current_state}}
Desired Situation: {{desired_state}}
Existing Requirements: {{requirements}}

Only output valid JSON.
{{~/user}}
{{#assistant~}}
{{gen "response" temperature=0.1}}
{{~/assistant}}

\end{Verbatim}

We now run this program with the values supplied by the user as well as the
current set of requirements already present in the database.

\begin{minted}{python}
prog = self.requalizer(
	product=self.data["product"],
	vision=self.data["vision"],
	niche=self.data["niche"],
	current_state=self.data["current_state"],
	desired_state=self.data["desired_state"],
	requirements=self.data["requirements"],
	count=count
)
\end{minted}

The example requirement that we would get from this program will look like this
when converted from JSON to YAML:

\begin{Verbatim}[tabsize=2]
name: Language model integration
description: Integrate state-of-the-art language models to automate the scrum process
customer_type: Entrepreneurial and innovative company
stakeholders: Product Managers, Scrum Masters, Developers
importance: 5
assumptions: Assuming that the company is willing to invest in technology to boost
  productivity
risks: Possible risks include the need for technical expertise to integrate the language
  models
\end{Verbatim}

We can add as many fields as we want and the model will try to compute the most
relevant values for each field.

%% file: parts/method-product-features.tex
Product features are properties of the product that enable the product to
fulfill the requirements necessary in order to be a good tool that can be used
by a niche participant to go from their current state to their desired state.

In simple terms, product features fulfill customer requirements and customer
requirements are based on the customer's desire to go from current state to
their desired state.

We can thus identify product features entirely from the requirements in the
previous state. Our hyper-parameters for this step become:

\begin{itemize}
	\item \textbf{Product}: the name of the product we are building.
	\item \textbf{Product Vision}: our vision for the product (this is what WE
		want the product to be).
	\item \textbf{Requirements}: a list of requirements as descriptions (we take
		the JSON field generated in the previous step and make a list of
		descriptions).
	\item \textbf{Existing Features}: since the process is iterative, we have to
		include features we have already identified earlier in order to avoid
		the model generating duplicate features.
	\item \textbf{Avoid Features}: this is a list of descriptions of features we
		want to avoid. For example: "avoid any feature that involves Ruby
		programming because this product is not written in Ruby".
\end{itemize}

We write a Guidance program to generate features using GPT-3.5. This program
will likely work even better on GPT-4. However, it is a good practice to make
your program so good that it works on GPT-3 because GPT-3 is much less smart at
figuring out what you mean when you instruct it. You have to be explicit.

\begin{Verbatim}
{{#system~}}
You are a helpful assistant that helps me with Scrum project planning.

Your responses contain only valid JSON.

Your task today is to generate product features based on product vision and product requirements.

I will provide you with the following information:
Product: name of the product we are building
Product Vision: a description of the product vision
Requirements: a list of product requirements
Story Features: a list of features identified during story mapping
Existing Features: features that are already identified and should not be repeated
Avoid Features: a list of feature types that you should avoid

Make sure that:
1) All features that you generate must fit together with the product vision.
2) Each feature must be unique and separate from all other features in the existing features list.
3) Do not repeat yourself in feature titles. Make each feature unique and well defined.
4) Prioritize creating new features based on story features which do not already exist in the
   existing features list
5) Do not repeat existing features or features that have similar functionality.
6) Avoid mentioning the features in the "Avoid Features" list in your output.
7) Make sure that each feature you generate meets at least one requirement.
8) Generate only {{count}} features at a time.

The features should be returned as a JSON list in the following format.

RESPONSE FORMAT:
[
  {
    "name": "Title of the feature",
    "explanation": "explanation of the feature",
    "reasoning": "Reasoning for first feature",
    "goal": "a concise name of product goal met by the feature"
  },
  {
    "name": "Another feature",
    "explanation": "explanation of the second feature",
    "reasoning": "Reasoning for why we should develop this feature",
    "goal": "name of product goal"
  }
]
{{~/system}}
{{#user~}}
Product: {{product}}
Product Vision: {{vision}}
Requirements: {{requirements}}
Story Features: {{story_features}}
Existing Features: {{features}}
Avoid Features: {{avoid}}

Only output valid JSON.
{{~/user}}
{{#assistant~}}
{{gen "response" temperature=0.1}}
{{~/assistant}}

\end{Verbatim}

We can run this program with the following parameters:

\begin{minted}{python}
prog = featurizer(
	product=self.data["product"],
	vision=self.data["vision"],
	requirements=[req["description"] for req in self.data["requirements"]],
	features = [feature["name"] for feature in self.data["features"]],
	avoid=self.data["avoid"],
	count=count
)
\end{minted}

Running this program on GPT-3.5-turbo returns the following list (the program
generates JSON data but I have converted it to YAML here for easy word wrapping:

\begin{Verbatim}[tabsize=2]
---
- name: Automated Product Vision Generation
  explanation: Automatically generate a product vision from user inputted product
    features and requirements using large language models.
  reasoning: This feature will allow users to quickly and easily generate a product
    vision without having to spend time manually writing it out. It will also ensure
    that the product vision is aligned with the product features and requirements.
- name: Automated User Story Generation
  explanation: Automatically generate user stories from product features using large
    language models.
  reasoning: This feature will save time for the product owner by automatically generating
    user stories from the product features. It will also ensure that the user stories
    are aligned with the product features.
- name: Automated Acceptance Criteria Generation
  explanation: Automatically generate acceptance criteria for user stories using large
    language models.
  reasoning: This feature will save time for the developers by automatically generating
    acceptance criteria for user stories. It will also ensure that the acceptance
    criteria are aligned with the user stories.
\end{Verbatim}

It is beneficial to generate three features at a time and then provide the model
with only feature titles in the list of existing features. The model is good at
avoiding features that have already been specified allowing us to generate
features iteratively.

%% file: parts/method-story-mapping.tex
Once we have all the features and we are satisfied with our list, we can start
working on user stories. We will use a new guidance program that we write
specifically for development of user stories.

\begin{Verbatim}
{{~! User stories identification from requirements }}
{{#system~}}
You are a helpful Scrum assistant helping me with identifying user stories from user requirements.

I will provide you with the following information:
Product: the name of the product
Niche: target niche of the product
Product Vision: a description of the product vision
Existing Features: a list of already identified features
Current State: user's current situation
Desired State: user's desired situation
Existing stories: a list of existing user story titles for stories that should not be included in
your response.

You should only include stories in your output that are relevant for helping user go from their
current state to their desired state and which have not been included in the existing stories list.

Generate user stories that capture how the user will be using the product to go from their current
state to their desired state. These user stories should be small enough to be completed within a
sprint and should be written from the perspective of the stakeholder who is interested in the story.

If a new user story requires a new feature that has not been identified yet then use the name of the
new feature instead as the "feature" field of the story.

Your response should be a list of user stories that are connected to features.

RESPONSE FORMAT:
[
  {
    "name": "User story title",
    "feature": "Name of the feature that would be required for this user story",
    "epic": "Suggest an epic name for grouping stories like this one",
    "reasoning": "Reasoning for including this user story"
  }
  {
    "name": "As a [user role], I want [goal/desire] so that [benefit]"
    "feature": "Description of the feature that this user story would require",
    "epic": "Epic to which this story belongs",
    "reasoning": "Why did you include this story"
  }
]

Make sure user story titles are unique and different. Do not repeat yourself and do not include
already existing stories. Consider multiple types of users including developers. Use name format
like "As a [user role], I want [goal/desire] so that [benefit]".

Generate {{count}} new user stories.
{{~/user}}
{{#user~}}
Product: {{product}}
Niche: {{niche}}
Product Vision: {{vision}}
Existing Features: {{features}}
Current State: {{current_state}}
Desired State: {{desired_state}}
Existing stories: {{stories}}
{{~/user}}
{{#assistant~}}
{{gen "response" temperature=0.1}}
{{~/assistant}}
\end{Verbatim}

We now run the program from python.

\begin{minted}{python}
prog = program(
	product=self.data["product"],
	vision=self.data["vision"],
	features = [feature["name"] for feature in self.data["features"]],
	goals=[feature["goal"] for feature in self.data["features"]],
	stories = [story["name"] for story in self.data["stories"]],
	count=count
)
\end{minted}

This program will generate the following information (JSON converted to YAML):

\begin{Verbatim}[tabsize=2]
---
- name: As a product owner, I want to be able to input product features and have the
    system automatically generate a prioritized product backlog
  feature: Automated Product Backlog Generation
  reasoning: This story addresses the goal of automating the process of generating
    a product backlog.
- name: As a scrum master, I want to be able to generate a sprint review report with
    just a few clicks, so that I can save time and focus on other important tasks
  feature: Automated Sprint Review Report Generation
  reasoning: This story addresses the goal of automating the process of generating
    a sprint review report.
- name: As a development team member, I want to be able to schedule a sprint planning
    meeting with ease, so that I can focus on preparing for the meeting
  feature: Automated Sprint Planning Meeting Scheduler
  reasoning: This story addresses the goal of automating the process of scheduling
    a sprint planning meeting.
\end{Verbatim}

Note that the second story that has been generated says "with just a few
clicks". This is interesting because we have not told the model whether we are
building a command line application or a graphical one. It just assumed that we
are building a graphical product. This is one of the drawbacks of generalized
language models - they are not trained on your domain data and so the output
heavily depends on what information you provide.

Furthermore, the information we can provide is limited because we are limited by
the maximum prompt size. Therefore in situations like this where we are using
zero shot learning (just asking the model without any fine tuning) we inevitably
get wrong formulations and even wrong stories sometimes. This problem can be
solved with either fine tuning or simply by pairing the model with human review.

%% file: parts/method-acceptance.tex
We can use the information we have so far to generate acceptance criteria for
all stories that do not have it yet.

In Scrum, acceptance criteria is always generated for user stories and are
neither for epics nor tasks. Acceptance criteria defines the specific conditions
that must be met for a user story to be completed.

While user stories represent the desired functionality or features from the
perspective of the end user, the acceptance criteria is used to clearly
articulate the requirements and expectations for each user story.

In order to automatically generate a suggested acceptance criteria for our user
stories, we can leverage the list of manually identified requirements as well as
the data in the user story.

We then run the acceptance criteria Guidance program for each user story.

\begin{Verbatim}
{{~! Acceptance criteria generation for user stories }}
{{#system~}}
You are a helpful assistant that helps me write acceptance criteria for SCRUM user stories.

I will provide you with the following information:
Product: the product to which the story relates
Story: A particular user story for which you will generate acceptance criteria
Requirements: a full set of identified user requirements

You must follow the following criteria:
1) You must output the result as json
2) You must be specific
3) You must only consider requirements that are relevant for the particular user story

You should only respond with a Python list in JSON format that can be parsed using json.loads:
RESPONSE FORMAT:
[
	"list of specific conditions or requirements that need to be met for the story to be considered complete",
	"description or summary of acceptance criteria item",
	"third acceptance criteria",
	...
]

Generate {{count}} acceptance criteria.
{{~/system}}
{{#user~}}
Product: {{product}}
Story: {{story}}
Requirements: {{requirements}}
{{~/user}}
{{#assistant~}}
{{gen "response" temperature=0 max_tokens=300}}
{{~/assistant}}

\end{Verbatim}

The python code is very similar each time so it can be omitted. We are basically
running the above guidance program with our own parameters against GPT-3.5.

The result is the following updated user story:
\begin{Verbatim}[tabsize=2]
---
name: As a scrum master, I want to be able to generate a sprint review report with
  just a few clicks, so that I can save time and focus on other important tasks
feature: Automated Sprint Review Report Generation
reasoning: This story addresses the goal of automating the process of generating a
  sprint review report.
acceptance:
- The sprint review report should be generated automatically with just a few clicks
- The report should include a summary of completed user stories, including their acceptance
  criteria and any relevant metrics
- The report should also include a list of unfinished user stories and their current
  status, as well as any impediments that were encountered during the sprint
\end{Verbatim}

%% file: parts/method-task-decomposition.tex
User stories now need to be decomposed into tasks.

We will now write a language program to generate tasks and subtasks given
the information we have generated so far.

The task generation program takes the user story as input and outputs a list of
tasks with subtasks.

\begin{Verbatim}
{{~! Breaking down user stories into tasks and subtasks }}
{{#system~}}
You are a helpful assistant helping with with python coding. Today your job is to help me define
tasks for user stories. 

I will give you the following information:
name: name of the user story
feature: the product feature that this story implements
reasoning: why this story is needed
acceptance: the acceptance criteria for the user story

Your job is to help me identify tasks for this user story.  There can be any number of tasks.
Identify as many as possible but at most 7.

You should only respond with JSON list of tasks with subtasks. 
RESPONSE FORMAT:
[
{
"task": "task title",
"subtasks": [list of subtasks]
},
{
"task": "second task title",
"subtasks": [list of subtasks]
}
...
]

Do not reply with anything else besides valid JSON data.
{{~/system}}
{{#user~}}
Story: {{story}}
{{~/user}}
{{#assistant~}}
{{gen "response" temperature=0.1}}
{{~/assistant}}
\end{Verbatim}

We can now run this program on all of our user stories that have an acceptance
criteria, and the result is that we get user stories with tasks and subtasks:

\begin{Verbatim}[tabsize=2]
name: As a product owner, I want to be able to customize the automated product backlog
  generation process by adding or removing features, so that I can tailor the backlog
  to my specific needs
feature: Automated Product Backlog Generation
reasoning: This user story allows the product owner to have more control over the
  generated product backlog, which can improve the overall planning process.
acceptance:
- The product owner should be able to add or remove features from the automated product
  backlog generation process
- The automated product backlog generation process should be able to generate product
  goals from the product features and requirements
- The automated product backlog generation process should be able to generate user
  stories from the product features, and break them down into tasks automatically
tasks:
- task: Create a user interface for customizing the automated product backlog generation
    process
  subtasks:
  - Design the user interface for adding or removing features
  - Implement the user interface using a suitable framework
  - Integrate the user interface with the automated product backlog generation process
- task: Modify the automated product backlog generation process to allow for customization
  subtasks:
  - Identify the appropriate points in the code where customization can be added
  - Implement the ability to add or remove features from the backlog generation process
  - Test the modified process to ensure that it still generates product goals and
    user stories correctly
\end{Verbatim}

%% file: parts/method-task-completion.tex
The tasks that we have now generated correspond to the detailed instructions
that must be executed in order to complete the user story. This is the point
where we can greatly enhance the user experience further by taking the task that
needs to be done and providing additional assistance in the form of domain
specific knowledge and information that can be used to complete the task.

The whole process of task completion itself can and will be automated, but we
consider it more useful to invite domain experts and actual teams to work on the
tasks because the results of domain specific expertise is almost certainly going
to be much better than either machine or a human working on their own.

The question we ask is "What is now needed in order to fully accomplish a
specific user story?". We must take into consideration the user story, the
product requirements and the product vision and then we generate questions and
search terms that the user can use in order to find the information for solving
the task.

This process is particularly effective when you use the output of this step to
search a domain specific vector database where you have stored specific
knowledge in the form of vector embeddings and where you can retrieve the right
knowledge based on similarity with the search query. However, this process works
just as well if you execute it manually by using search engines and existing
databases.

\begin{Verbatim}
{{~! Breaking down user stories into tasks and subtasks }}
{{#system~}}
You are a helpful assistant helping with with python coding. Today your job is to help me define
tasks for user stories. 

I will give you the following information:
name: name of the user story
feature: the product feature that this story implements
reasoning: why this story is needed
acceptance: the acceptance criteria for the user story

Your job is to help me identify tasks for this user story.  There can be any number of tasks.
Identify as many as possible but at most 7.

You should only respond with JSON list of tasks with subtasks. 
RESPONSE FORMAT:
[
{
"task": "task title",
"subtasks": [list of subtasks]
},
{
"task": "second task title",
"subtasks": [list of subtasks]
}
...
]

Do not reply with anything else besides valid JSON data.
{{~/system}}
{{#user~}}
Story: {{story}}
{{~/user}}
{{#assistant~}}
{{gen "response" temperature=0.1}}
{{~/assistant}}
\end{Verbatim}

The purpose of this program is to provide us with searchable terms that we can
use to search an internal database for answers. We can train the model using
in-prompt examples to respond with highly relevant terms. The output of this
program is a list of questions and search terms:

\begin{Verbatim}[tabsize=2]
resources:
- question: What programming language will the tool be developed in?
  concept: tool development
- question: What format will the user stories and priorities be imported in?
  concept: user story and priority import
- question: What historical data will be used for effort estimation?
  concept: historical data for effort estimation
- question: What is the team capacity and how will it be imported?
  concept: team capacity import
- question: What is the expected output format for the prioritized product backlog
    and sprint backlog?
  concept: output format for product backlog and sprint backlog
\end{Verbatim}

If you have ever had to resort to searching Google for answers, you probably
already know that the biggest problem is knowing what to search for. Language
models can help us bridge this gap by directly providing suggestions. This has
the potential of significantly reducing the time that developers need to spend
trying to find the information that they need in order to complete a task.

Building the knowledge database should be of the highest concerns for a company
because when you can easily go to the right document that describes the right
procedure then executing each task that has been added to the backlog is easy.
This is why maintaining good documentation for internal use is very important
especially when language models are integrated into the daily workflow.

This kind of question asking is a good example of how language models can be
used for directing your attention to places where it matters. A language model
can suggest that you read a book or research a paper that is highly relevant for
your work. It can also generate search terms that can be used in combination
with vector embeddings to find the materials that are most similar to the query.

%% file: parts/method-shortcut-plan.tex
There is a slightly different approach that we can also consider. In this
approach, instead of using Scrum, we simply use the current and the desired
state and have the language model continuously generate the next most optimal
task to take us from A to B.

The language program for this approach looks like this:

\begin{Verbatim}
{{#system~}}
You are a dedicated mentor, instructing me on the most important next task I should do that would
help me go from my current situation to my desired situation.

I will provide you with the following information:
Current Situation: a set of variables representing current situation
Desired Situation: a set of variables representing desired situation
Completed Tasks: a list of tasks that have already been completed along the path
Unattainable Tasks: Tasks that have proven too challenging to achieve so far.

You must follow the following criteria:
1) You should act as a mentor and guide me to the next task based on my current situation and
   progress towards desired state.
2) Please be very specific when describing the next task I should do.
3) The next task should follow a concise format, such as "Study [scientific paper]", "Read [book
   title] by [author]", "Code [software component]". It should be a single phrase. Do not propose
   multiple tasks at the same time. Do not mention anything else besides the best task I should do
   right now.
4) The next task should be novel and interesting. It should be highly relevant for attainment of the
   desired state if desired state has been achieved, it should promote surpassing it.
5) You may sometimes need to repeat some tasks in order to move forward. Only repeat previous tasks
   if absolutely necessary.
6) You must set task status to "done" if the desired state has been achieved. Do not set "status" to
   "progress" if the desired state has been clearly achieved. This is important.

You should respond in valid JSON list that can be parsed using json.loads. Do not write any other
text. Only write json according to the response structure below.

EXAMPLE INPUT:
Current State: {"apples": 5}
Desired State: {"apples": 10}
Tasks Completed: []

CORRECT RESPONSE:
[
  {
    "reasoning": "As you only have 5 apples and you need 10, the fastest way to get 10 apples is
    to buy 5 more",
    "task": "Buy 5 more apples",
    "status": "progress"
  }
]

EXAMPLE INPUT:
Current State: {"apples": 10}
Desired State: {"apples": 10}
Tasks Completed: ["Buy 5 more apples"]

CORRECT RESPONSE:
[
  {
    "reasoning": "As you have reached your current state there is not much more for you to do",
    "task": "Eat an apple",
    "status": "done"
  }
]
(here the status MUST be set to "done" because we have already reached our desired state)

EXAMPLE INPUT:
Current State: {"lemons": 15}
Desired State: {"lemons": 10}
Tasks Completed: ["Buy 10 more lemons"]

INCORRECT RESPONSE:
[
  {
    "reasoning": "As you have reached your current state there is not much more for you to do",
    "task": "Bake a lemon pie",
    "status": "progress"
  }
]
(here the status MUST be set to "done" because we have already reached our desired state)

RESPONSE FORMAT:
[
  {
    "reasoning": "In light of the data I have provided above, offer a comprehensive
       explanation for what my next task should be and why",
    "task": "Specify the next single task that requires my attention. Only one task at a time."
    "state": "'done' if desired state achieved, otherwise 'progress'"
  }
]

Make sure that if the desired state has been reached then "status" should be set to "done".

Generate {{count}} tasks as a json list object.
{{~/system}}
{{#user~}}
Current Situation: {{current_state}}
Desired Situation: {{desired_state}}
Completed Tasks: {{plan}}
{{~/user}}
{{#assistant~}}
{{gen "response" temperature=1.2}}
{{~/assistant}}

\end{Verbatim}

This approach generates a shortcut plan for achieving our goal. Let's try it out
with a sample task of renaming and sorting JPEG files in a directory based on
JPEG timestamp.

Input data to autoscrum:

\begin{Verbatim}
---
product: JPEG sorting script
vision: Easy command that can rename jpeg files in a directory to filenames that correspond
  date and time when photo was taken based on jpeg metadata
niche: Linux users
current_state:
  files_in_directory: unsorted with random names
desired_state:
  files_in_directory: renamed to jpeg timestamp and sorted by date photo taken
\end{Verbatim}

If we run this for three iterations without updating the desired state, we get a
plan that explains to us what to do in order to achieve the result. If we then
update the current state to "all files have names that correspond to their JPEG
timestamps" after three iterations, we can see that the model now returns status
"done".

\begin{Verbatim}
---
plan:
- reasoning: As the current situation is disorganized with random names attached to
    each image, we need to rename the files with their corresponding jpeg timestamp.
    Installing ExifTool to extract the timestamp would be the best way to perform
    this task. We would then need to sort the files according to their photo taken
    date. We could achieve this through a script and tagging each photo properly.
  task: Install ExifTool and write a script to rename files based on jpeg timestamp,
    then sort files by their photo taken date.
  status: progress
- reasoning: Since you already have a script to rename the photos and sort them by
    date taken, the next task should be to run the script on all the files in the
    designated directory.
  task: Run the script on all the files in your unsorted directory.
  status: progress
- reasoning: Now that you have renamed and sorted your photo files, the next step
    is to review the sorted files and delete any duplicates or unwanted photos.
  task: Review and delete any duplicates or unwanted photos.
  status: progress
- reasoning: Since all the photos already have timestamp filenames and are sorted
    by photo date, there is nothing more to do.
  task: Review your sorted photo collection and select your favorites
  status: done
\end{Verbatim}

One thing to note here is that the GPT-3.5 model is prone to continue generating
tasks even past the done point. It was necessary to set the current state to
exactly the same values as the desired state for the model to output "done"
right away.

The challenge here is that we have told the model to continue generating novel
and interesting tasks even if the task was done but GPT-3.5 still seems to
struggle with making sure that the "status" is always set to "done" when desired
state has been reached. GPT-4 does not struggle with this and seems to be
consistently setting the status to done once the desired state was clearly
reached.

%% file: parts/results-overview.tex
The GPT-3.5-turbo model has been observed to be extremely good at mapping out
user stories and making sure that these user stories are formulated according to
the desired JSON format. The reliability of the language model at returning
relevant and correct JSON has been almost flawless.

In order to make this work, the prompt (the Guidance programs) had to be
iteratively developed and extended with clear and detailed instructions for the
model to follow.

\begin{itemize}
	\item \textbf{Current And Desired Situation}: parameterizing on current and
		desired situation and letting the model bridge the way in between is one
		of the most exciting results of this study.
	\item \textbf{Effect Of Temperature}: increasing the temperature parameter
		seems to make the model suggest bolder tasks while lower value seems to
		suggest a more step by step approach.
\end{itemize}

%% file: parts/results-scrum.tex
Parameterizing the generation process on user's current and desired situation has
been an innovative approach in this paper and the GPT-3.5 model was very good at
brainstorming tasks that would take the user from their current to their desired
state.

We can demonstrate this using a simple example.

Suppose we have a barn with 10 chickens in it. We would like to have 200
chickens in it. We can start the generation process using a simple JSON file
with the following fields (we use YAML just for clean presentation):

\begin{Verbatim}[tabsize=2]
product: Happy Farm
vision: A farm full of chickens
niche: Farmers
current_state:
  chicken_count: 10
desired_state:
  chicken_count: 200
requirements: []
sprint_duration: 2 weeks
features: []
stories: []
avoid: []
\end{Verbatim}

We call the autoscrum script from the Github repository repeatedly to run all
steps of the process. Most of the time we opt for generating three items at a
time.

\begin{Verbatim}
autoscrum -f examples/happy-farm.json requirements -n 3
autoscrum -f examples/happy-farm.json requirements -n 3
autoscrum -f examples/happy-farm.json stories -n 3
autoscrum -f examples/happy-farm.json features -n 3
autoscrum -f examples/happy-farm.json acceptance -n 3
autoscrum -f examples/happy-farm.json tasks -n 3
autoscrum -f examples/happy-farm.json clarify -n 3
autoscrum -f examples/happy-farm.json plan -n 1
\end{Verbatim}

We can now run the requirements program and instruct it to generate three
requirements based on the above data (we can run it again to generate more
requirements).

\begin{Verbatim}[tabsize=2]
requirements:
- name: Increase chicken coop size
  description: The chicken coop needs to be expanded to accommodate more chickens.
  customer_type: Farmer
  stakeholders: User
  importance: 5
  assumptions: The farmer wants to increase the number of chickens they have.
  risks: The cost of expanding the chicken coop may be too high for the farmer.
- name: Purchase additional chickens
  description: The farmer needs to purchase more chickens to increase their flock.
  customer_type: Farmer
  stakeholders: User
  importance: 4
  assumptions: The farmer has the resources to purchase more chickens.
  risks: The new chickens may introduce disease to the existing flock.
- name: Implement a feeding and care schedule
  description: The farmer needs to implement a feeding and care schedule to ensure
    the health and productivity of the chickens.
  customer_type: Farmer
  stakeholders: User
  importance: 3
  assumptions: The farmer wants to maintain the health and productivity of their chickens.
  risks: The farmer may not have the time or resources to implement a feeding and
    care schedule.
\end{Verbatim}

What's fascinating here is that even though GPT-3.5 is a much less advanced
model than GPT-4, it still manages to highlight the risk that buying more
chickens could bring disease into the flock.

We can now continue and generate user stories by using the story Guidance
program.

\begin{Verbatim}[tabsize=2]
- name: As a farmer, I want to purchase more chickens easily so that I can increase
    my chicken count
  feature: Add a 'Buy Chickens' button to the user interface
  epic: Chicken Management
  reasoning: This story will allow the farmer to easily purchase more chickens and
    increase their chicken count
- name: As a farmer, I want to track the health of my chickens so that I can ensure
    they are healthy and productive
  feature: Add a 'Health Tracker' feature to the user interface
  epic: Chicken Management
  reasoning: This story will allow the farmer to monitor the health of their chickens
    and take necessary actions to keep them healthy and productive
- name: As a developer, I want to optimize the chicken feeding process so that the
    chickens grow faster and healthier
  feature: Add an 'Optimized Feeding Algorithm' feature to the backend
  epic: Chicken Management
  reasoning: This story will allow the developer to optimize the feeding process and
    improve the health and growth of the chickens
\end{Verbatim}

Here, the GPT-3.5 model believes that a farm is an application. The reason for
this could be somewhere in the prompt or it could be that the model is simply
biased towards software development. At this point you can rewrite the "Add
Chickens" button to something like "Place an order for more chickens". Or
perhaps we should actually provide the farmer with the "Buy Chickens" button and
market it as an application? After all, we had the "product" name set to "Happy
Farm". That could be an application name.

Language models like GPT-3.5 and beyond display incredible ability of guiding
the human thought process. This is extremely valuable and it goes far beyond
just parroting some knowledge (as some people might think).

The second story above tells us that we need to have a health tracker for the
chickens. That would definitely make sure that the farmer knows if his chickens
ever get sick.

The third story describes the chicken feeding process. Obviously this process
should be automated, especially if the farmer hopes to have 200 healthy
chickens.

Just by generating the user stories we have managed to capture the key actions
necessary to go from 10 chickens to 200. In AutoScrum Github demo you
can run this process many times and generate as many user stories as you like.

Next let's generate features.

\begin{Verbatim}[tabsize=2]
- name: Expand Chicken Coop
  explanation: The chicken coop needs to be expanded to accommodate more chickens.
  reasoning: Expanding the chicken coop will allow the farmer to increase their flock
    and ultimately increase productivity and profits.
  goal: Increase chicken flock size
- name: Buy More Chickens
  explanation: The farmer needs to purchase more chickens to increase their flock.
  reasoning: Increasing the number of chickens will increase productivity and profits.
  goal: Increase chicken flock size
- name: Implement Feeding and Care Schedule
  explanation: The farmer needs to implement a feeding and care schedule to ensure
    the health and productivity of the chickens.
  reasoning: A feeding and care schedule will ensure that the chickens are healthy
    and productive, which will ultimately increase profits.
  goal: Ensure chicken health and productivity
\end{Verbatim}

In the current implementation, the idea is that you would be generating your
product features in parallel with identifying user stories. So while user
stories have a "feature" field in them, the product features are designed to be
stored as a separate list.

Let's create an acceptance criteria for our stories. We can now expand our list
of stories with acceptance criteria for each story. I will not edit the original
generated user stories before generating acceptance criteria, but technically
you both could and should do that because during the story refinement process
you would typically both delete unneeded stories and rephrase the existing
ones.

\begin{Verbatim}[tabsize=2]
- name: As a farmer, I want to purchase more chickens easily so that I can increase
    my chicken count
  feature: Add a 'Buy Chickens' button to the user interface
  epic: Chicken Management
  reasoning: This story will allow the farmer to easily purchase more chickens and
    increase their chicken count
  acceptance:
  - The 'Buy Chickens' button should be prominently displayed on the user interface
  - The 'Buy Chickens' button should be easy to find and use
  - The 'Buy Chickens' button should allow the farmer to select the number of chickens
    they want to purchase and complete the transaction smoothly
- name: As a farmer, I want to track the health of my chickens so that I can ensure
    they are healthy and productive
  feature: Add a 'Health Tracker' feature to the user interface
  epic: Chicken Management
  reasoning: This story will allow the farmer to monitor the health of their chickens
    and take necessary actions to keep them healthy and productive
  acceptance:
  - The 'Health Tracker' feature should allow the farmer to input and track the number
    of chickens in their flock
  - The 'Health Tracker' feature should allow the farmer to input and track the feeding
    and care schedule for each chicken
  - The 'Health Tracker' feature should provide alerts or notifications to the farmer
    if any chicken's health or productivity falls below a certain threshold
- name: As a developer, I want to optimize the chicken feeding process so that the
    chickens grow faster and healthier
  feature: Add an 'Optimized Feeding Algorithm' feature to the backend
  epic: Chicken Management
  reasoning: This story will allow the developer to optimize the feeding process and
    improve the health and growth of the chickens
  acceptance:
  - The optimized feeding algorithm should take into account the age, weight, and
    breed of the chickens to determine the appropriate amount and type of feed to
    be dispensed.
  - The algorithm should also consider the number of chickens in the coop and adjust
    the feeding schedule accordingly.
  - The farmer should be able to view the feeding schedule and make adjustments as
    needed to ensure the health and productivity of the chickens.
\end{Verbatim}

Once the acceptance criteria has been generated, we can now generate tasks for
each user story. While we are at it, let's also generate the questions as well.

Only the second story is included here as demonstration, but you can find the
full results from this run in the Github repository (\url{https:
//github.com/autoscrum/autoscrum}).

\begin{Verbatim}[tabsize=2]
name: As a farmer, I want to track the health of my chickens so that I can ensure
  they are healthy and productive
feature: Add a 'Health Tracker' feature to the user interface
epic: Chicken Management
reasoning: This story will allow the farmer to monitor the health of their chickens
  and take necessary actions to keep them healthy and productive
acceptance:
- The 'Health Tracker' feature should allow the farmer to input and track the number
  of chickens in their flock
- The 'Health Tracker' feature should allow the farmer to input and track the feeding
  and care schedule for each chicken
- The 'Health Tracker' feature should provide alerts or notifications to the farmer
  if any chicken's health or productivity falls below a certain threshold
tasks:
- task: Create a database schema for storing chicken health data
  subtasks:
  - Define the necessary tables and columns for storing chicken health data
  - Determine the appropriate data types for each column
  - Set up relationships between tables as necessary
- task: Implement the 'Health Tracker' feature in the user interface
  subtasks:
  - Add a new section to the user interface for the 'Health Tracker' feature
  - Create forms for inputting and tracking the number of chickens in the flock
  - Create forms for inputting and tracking the feeding and care schedule for each
    chicken
  - Implement a notification system for alerting the farmer if any chicken's health
    or productivity falls below a certain threshold
- task: Integrate the 'Health Tracker' feature with the database
  subtasks:
  - Ensure that data entered through the user interface is properly stored in the
    database
  - Ensure that data retrieved from the database is properly displayed in the user
    interface
  - Implement data validation to prevent incorrect or invalid data from being stored
    in the database
- task: Test the 'Health Tracker' feature
  subtasks:
  - Create test cases for each feature of the 'Health Tracker'
  - Manually test the 'Health Tracker' feature to ensure it is functioning as expected
  - Automate tests for the 'Health Tracker' feature to ensure it continues to function
    properly in the future
resources:
- question: What are the common health issues that chickens face?
  concept: chicken health
- question: What are the common signs of a healthy chicken?
  concept: chicken health
- question: What are the common signs of an unhealthy chicken?
  concept: chicken health
- question: What are the common feeding and care schedules for chickens?
  concept: chicken feeding and care
- question: What are the best practices for setting up a notification system?
  concept: notification system
\end{Verbatim}

The questions are particularly interesting here because it's very obvious that
if you are implementing a user story that is connected to tracking chicken
health, you should definitely know what signs you should be tracking.

%% file: parts/results-shortcut.tex
This is a plan that was iteratively developed by this program for going from 10
chickens to 200 chickens. This plan does not use scrum. In fact, it does not do
anything besides looking at the current state, desired state and currently
completed tasks.

The plan is run for 10 steps and culminates with LLM considering the task done
and outputting "status": "done". While running this program we also iteratively
update the chicken count so that the system is always working from the most up
to date current state.

\begin{Verbatim}
plan:
- reasoning: To increase the chicken count, we need to acquire more chickens. One
    way to do this is to purchase them from a local farm or hatchery.
  task: Buy 50 baby chicks from a local farm or hatchery.
  status: progress
- reasoning: To continue increasing the chicken count, it's necessary to provide a
    suitable environment for the chicks to grow and thrive.
  task: Build a chicken coop that can accommodate at least 100 chickens.
  status: progress
- reasoning: To increase the chicken count, it's necessary to purchase more baby chicks
    and raise them to maturity.
  task: Buy 50 more baby chicks from a local farm or hatchery.
  status: progress
- reasoning: You have already bought 100 baby chicks and built a coop that can accommodate
    100 chickens. The next task should be to buy another 50 baby chicks to reach your
    desired chicken count of 200.
  task: Buy 50 more baby chicks from a local farm or hatchery.
  status: progress
- reasoning: You have already completed the initial steps of buying baby chicks and
    building a coop. The next step should be to ensure that the chickens are healthy
    and well-fed to increase their productivity.
  task: Consult with a veterinarian to develop a feeding and health plan for the chickens.
  status: progress
- reasoning: You have already completed several tasks to increase your chicken count,
    but you are still short of your desired count. The next task should be to purchase
    20 more baby chicks from a local farm or hatchery.
  task: Buy 20 more baby chicks from a local farm or hatchery.
  status: progress
- reasoning: You have made significant progress towards your desired situation of
    having 200 chickens. However, you are still short by 20 chickens. Therefore, the
    next task you should do is to buy 20 more baby chicks from a local farm or hatchery.
  task: Buy 20 more baby chicks from a local farm or hatchery.
  status: progress
- reasoning: Since you have already completed several tasks to increase your chicken
    count and have reached your desired state, it's time to focus on maintaining the
    health and productivity of your chickens.
  task: Implement a regular cleaning and maintenance schedule for the chicken coop
    and run.
  status: progress
- reasoning: Since you have already completed several tasks related to acquiring and
    taking care of chickens, the next task should be to start selling eggs or meat
    to generate income.
  task: Research local markets and determine the best way to sell your chicken products.
  status: progress
- reasoning: Since you have completed all the necessary tasks to reach your desired
    state, there is not much more for you to do.
  task: Collect eggs from the chickens and sell them in the local market.
  status: done
\end{Verbatim}

%% file: parts/results-temperature.tex
Throughout this paper, GPT-3.5-turbo model was used with temperature setting set
to 0-0.1.

Here is the output of the same chicken farm plan generated with temperature 1.2
instead of 0.

\begin{Verbatim}[tabsize=2]
- reasoning: In order to increase your chicken count quickly, it is best to purchase
    more chickens. Research nearby hatcheries and farm animal stores to find where
    you can purchase chicks.
  task: Buy 190 more chickens
  status: progress
- reasoning: Since you only need 10 more chickens to reach your desired state and
    you have already purchased 190, the best thing to do is to buy the remaining 10
    chickens and add them to your livestock.
  task: Buy 10 more chickens
  status: progress
- reasoning: As you have already bought all 200 chickens needed to reach the desired
    situation, your focus should be now on taking care of them.
  task: Build a coop for the chickens
  status: progress
- reasoning: Now that your chicken count is at 200, it is important to take care of
    them so that you can reap the benefits of having a healthy flock. Cleaning the
    coop regularly is a crucial aspect of this.
  task: Clean the chicken coop
  status: progress
- reasoning: 'Now that you have the chickens and their coop is built and clean, the
    next task should be to form a feeding plan. Proper nutrition determines egg production,
    chicken growth and overall health. '
  task: Create a feeding plan for your chickens
  status: progress
- reasoning: Since you have already achieved your desired state of having 200 chickens,
    there is not much left to do. However, in order to maintain good health and egg
    production of your chickens, it is important to keep track of their health on
    a regular basis.
  task: Schedule a routine checkup with a veterinarian for your chickens
  status: progress
- reasoning: Since you have completed all the necessary steps for taking care of your
    chickens and also achieved your desired state of 200 chickens, I believe the best
    next task for you would be to monitor their growth, health, and productivity regularly.
  task: Monitor chicken growth, health, and productivity regularly.
  status: done
\end{Verbatim}

This plan is much more aggressive at reaching the goal but then spends
considerable time stabilizing the situation afterwards.

%% file: parts/final-thoughts.tex
This paper demonstrates the extraordinary ability of the ChatGPT transformer
model to assist us with project planning. When we combine the ability of ChatGPT
with human expertise, we get the best of both worlds. ChatGPT can sometimes
suggest outrageous ideas and humans can be there to consider which ideas are
important and make sense to implement.

The iterative nature of the process makes adjustments very easy at every step of
the way. There are lots of opportunities for improvement and this is just the
beginning of a new "industrial revolution" where language models assist us in
everything we do.

If you feel like sharing your input on this paper with me directly then you can
send me a message on LinkedIn or email me.

\begin{center}
	\url{https://www.linkedin.com/in/martinschroder/}
\end{center}
\begin{center}
	\url{martin.schroder@swedishembedded.com}
\end{center}

%% file: template.bbl
\begin{thebibliography}{9}
\providecommand{\natexlab}[1]{#1}
\providecommand{\url}[1]{\texttt{#1}}
\expandafter\ifx\csname urlstyle\endcsname\relax
  \providecommand{\doi}[1]{doi: #1}\else
  \providecommand{\doi}{doi: \begingroup \urlstyle{rm}\Url}\fi

\bibitem[Wang et~al.(2023{\natexlab{a}})Wang, Xie, Jiang, Mandlekar, Xiao, Zhu,
  Fan, and Anandkumar]{wang2023voyager}
Guanzhi Wang, Yuqi Xie, Yunfan Jiang, Ajay Mandlekar, Chaowei Xiao, Yuke Zhu,
  Linxi Fan, and Anima Anandkumar.
\newblock Voyager: An open-ended embodied agent with large language models,
  2023{\natexlab{a}}.

\bibitem[Xu et~al.(2023)Xu, Hong, Li, Hu, Chen, and Zhang]{xu2023tool}
Qiantong Xu, Fenglu Hong, Bo~Li, Changran Hu, Zhengyu Chen, and Jian Zhang.
\newblock On the tool manipulation capability of open-source large language
  models, 2023.

\bibitem[Bubeck et~al.(2023)Bubeck, Chandrasekaran, Eldan, Gehrke, Horvitz,
  Kamar, Lee, Lee, Li, Lundberg, Nori, Palangi, Ribeiro, and
  Zhang]{bubeck2023sparks}
Sébastien Bubeck, Varun Chandrasekaran, Ronen Eldan, Johannes Gehrke, Eric
  Horvitz, Ece Kamar, Peter Lee, Yin~Tat Lee, Yuanzhi Li, Scott Lundberg,
  Harsha Nori, Hamid Palangi, Marco~Tulio Ribeiro, and Yi~Zhang.
\newblock Sparks of artificial general intelligence: Early experiments with
  gpt-4, 2023.

\bibitem[Soong et~al.(2023)Soong, Sridhar, Si, Wagner, Sá, Yu, Karagoz, Guan,
  Hamadeh, and Higgs]{soong2023improving}
David Soong, Sriram Sridhar, Han Si, Jan-Samuel Wagner, Ana Caroline~Costa Sá,
  Christina~Y Yu, Kubra Karagoz, Meijian Guan, Hisham Hamadeh, and Brandon~W
  Higgs.
\newblock Improving accuracy of gpt-3/4 results on biomedical data using a
  retrieval-augmented language model, 2023.

\bibitem[Wang et~al.(2023{\natexlab{b}})Wang, Xu, Lan, Hu, Lan, Lee, and
  Lim]{wang2023planandsolve}
Lei Wang, Wanyu Xu, Yihuai Lan, Zhiqiang Hu, Yunshi Lan, Roy Ka-Wei Lee, and
  Ee-Peng Lim.
\newblock Plan-and-solve prompting: Improving zero-shot chain-of-thought
  reasoning by large language models, 2023{\natexlab{b}}.

\bibitem[Kang et~al.(2023)Kang, Laroche, Yuan, Trischler, Liu, and
  Fu]{kang2023think}
Jikun Kang, Romain Laroche, Xindi Yuan, Adam Trischler, Xue Liu, and Jie Fu.
\newblock Think before you act: Decision transformers with internal working
  memory, 2023.

\bibitem[Xie et~al.(2023)Xie, Xie, Lin, Wei, Li, Kong, Chen, Zhuo, Hu, and
  Li]{xie2023olagpt}
Yuanzhen Xie, Tao Xie, Mingxiong Lin, WenTao Wei, Chenglin Li, Beibei Kong, Lei
  Chen, Chengxiang Zhuo, Bo~Hu, and Zang Li.
\newblock Olagpt: Empowering llms with human-like problem-solving abilities,
  2023.

\bibitem[Zhu et~al.(2023)Zhu, Chen, Tian, Tao, Su, Yang, Huang, Li, Lu, Wang,
  Qiao, Zhang, and Dai]{zhu2023ghost}
Xizhou Zhu, Yuntao Chen, Hao Tian, Chenxin Tao, Weijie Su, Chenyu Yang, Gao
  Huang, Bin Li, Lewei Lu, Xiaogang Wang, Yu~Qiao, Zhaoxiang Zhang, and Jifeng
  Dai.
\newblock Ghost in the minecraft: Generally capable agents for open-world
  environments via large language models with text-based knowledge and memory,
  2023.

\bibitem[Liu et~al.(2023)Liu, He, Wang, Wang, Wang, Chen, Zhang, Lai, Yang, Li,
  Yu, Li, Chen, Yang, Zhu, Wang, Wang, Luo, Dai, and Qiao]{liu2023interngpt}
Zhaoyang Liu, Yinan He, Wenhai Wang, Weiyun Wang, Yi~Wang, Shoufa Chen,
  Qinglong Zhang, Zeqiang Lai, Yang Yang, Qingyun Li, Jiashuo Yu, Kunchang Li,
  Zhe Chen, Xue Yang, Xizhou Zhu, Yali Wang, Limin Wang, Ping Luo, Jifeng Dai,
  and Yu~Qiao.
\newblock Interngpt: Solving vision-centric tasks by interacting with chatgpt
  beyond language, 2023.

\end{thebibliography}
